\documentclass{jpsj3}
\usepackage{graphicx}
\usepackage{txfonts}
\usepackage{amsfonts}
\usepackage{bm}
\usepackage{euscript}
\usepackage{amssymb}

\title{Momentum-Space Renormalization Group Transformation in Bayesian Image Modeling by Gaussian Graphical Model}

\author{
\name{Kazuyuki Tanaka
\thanks{E-mail: kazu@smapip.is.tohoku.ac.jp}}$^{1}$, 
\name{Masamichi Nakamura}$^{1}$,
\name{Shun Kataoka}$^{1}$, 
\name{Masayuki Ohzeki}$^{1}$, 
\name{Muneki Yasuda}$^{2}$
}
\inst{
\address{$^{1}$Graduate School of Information Sciences,
Tohoku University, Sendai 980-8579, Japan}\\
\address{$^{2}$Graduate School of Science and Engineering, 
Yamagata University, 
Yonezawa 992-8510, Japan}\\
} %\\

\abst{
A new Bayesian modeling method is proposed 
by combining the maximization of the marginal likelihood 
with a momentum-space renormalization group transformation
for Gaussian graphical models.
Moreover, we present a scheme for computint the 
statistical averages of hyperparameters 
and mean square errors in our proposed method based on a momentum-space 
renormalization transformation.
}

\kword{statistical-mechanical informatics,
Markov random fields, renormalization group}

\begin{document}

\maketitle

Probabilistic graphical models based on Bayesian statistics 
are powerful tools 
for carrying out 
statistical inferences{\cite{KollerFriedman2009}
Probabilistic graphical models 
are regarded as an application of classical spin systems
from statistical-mechanical 
viewpoint{\cite{Nishimori2001,MezardMontanari2009}}.
Moreover, these provide  
many useful applications, not only for image processing 
but also for other inference systems
in high-dimensional data driven statistical 
approaches{\cite{KuwataniNagataOkadaToriumi2014A,
KuwataniNagataOkadaToriumi2014B,KataokaYasudaFurtlehnerTanaka2014}}.
The present authors previously 
proposed a novel and efficient method 
to accelerate hyperparameter estimations 
in Bayesian image segmentation problems
by introducing real-space renormalization 
group approaches{\cite{TanakaKataokaYasudaOhzeki2015}}.
Now we focus on another renormalization group technique, 
which is referred to 
as momentum-space renormalization group 
approach{\cite{AmitMartin-Mayor2005}}.
It is interesting to consider the momentum-space
renormalization group approaches for
accelerating the hyperparameter estimations in Bayesian modeling.

In the present paper, 
we propose a novel hyperparameter estimation scheme 
that combines momentum space renormalization group approaches
with the maximization of the marginal likelihood.
We introduce momentum space renormalization group approaches 
in Bayesian modeling in terms of Gaussian graphical models.

We introduce an $N{\times}N$ square grid graph 
$(V(N),E(N))$ 
where $V(N)=\{(x,y)|x=0,1,2,{\cdots},N-1,y=0,1,{\cdots},N-1\}$ 
is the set of all pixels
and $E(N)$ is the set of all edges 
consisting of pairs of pixels 
$\{(x,y),(x+1,y)\}$, $\{(x,y),(x,y+1)\}$,
$\{(x,y),(x+1,y+1)\}$ and $\{(x,y),(x+1,y-1)\}$.
The square grid graph has periodic boundary conditions 
along the abscissa and ordinate, respectively.
We define the state variables $f_{x,y}$ and $g_{x,y}$ 
($(x,y){\in}V(N)$),
which take any real values in the interval $(-{\infty},+{\infty})$. 
Using these state variables, we introduce 
the state vectors 
${\bm{f}}={\left(f_{x,y}{\big |}(x,y){\in}V(N)\right)}$
and 
${\bm{g}}={\left(g_{x,y}{\big |}(x,y){\in}V(N)\right)}$
as column vectors.
We consider a probability 
density function $P({\bm{f}}|{\alpha},{\gamma})$ 
and a conditional probability density function 
$P({\bm{g}}|{\bm{f}},{\beta})$ 
defined by 
\begin{eqnarray}
P({\bm{f}}|{\alpha},{\gamma}) 
&{\propto}&
{\exp}{\Bigg (}
   -{\frac{1}{2}}{\gamma}
    {\sum_{(x,y){\in}V(N)}}{f_{x,y}}^{2}
\nonumber\\
& &{\hspace{-1.50cm}}
   -{\frac{1}{2}}{\alpha}
    {\sum_{(x,y){\in}V(N)}}
    {\left({\left(f_{x,y}-f_{x+1,y}\right)}^{2}
          +{\left(f_{x,y}-f_{x,y+1}\right)}^{2}\right)}
      {\Bigg )},
\label{GaussianGraphicalModel-Prior}
\end{eqnarray}
\begin{eqnarray}
P({\bm{g}}|{\bm{f}},{\beta}) \equiv  
    {\prod_{(x,y){\in}V(N)}}
    {\sqrt{{\frac{{\beta}}{2{\pi}}}}}
        {\exp}{\Big (}-{\frac{1}{2}}{\beta}
        {\big (}f_{x,y}-g_{x,y}{\big )}^{2}{\Big )},
\label{GaussianGraphicalModel-DataDominant}
\end{eqnarray}
on the state space 
$(-{\infty},+{\infty})^{|V(N)|}$. 

Now we set a new positive integer $n$ with $n{\le}N$,
and introduce the following scale transformation 
from the state vectors ${\bm{f}}$ and ${\bm{g}}$
in the space $V(n)$ to new state vectors 
${\bm{f'}} = {\left(f'_{x,y}{\big |}(x,y){\in}V(n)\right)}$
and
${\bm{g'}} = {\left(g'_{x,y}{\big |}(x,y){\in}V(n)\right)}$:
\begin{eqnarray}
{\bm{f'}}
\equiv
{\left({\frac{n}{N}}\right)}^{{\psi}}
{\bm{U}}^{\dag}(n){\bm{B}}(n,N){\bm{U}}(N){\bm{f}},
\label{ScaleTransformationF}
\end{eqnarray}
\begin{eqnarray}
{\bm{g'}}
\equiv 
{\left({\frac{n}{N}}\right)}^{{\phi}}
{\bm{U}}^{\dag}(n){\bm{B}}(n,N){\bm{U}}(N){\bm{g}},
\label{ScaleTransformationG}
\end{eqnarray}
where
\begin{eqnarray}
{\langle}k',l'|{\bm{B}}(n,N)|k,l{\rangle}
\equiv
{\delta}_{k,k'}
{\delta}_{l,l'}
~{\left((k',l'){\in}W(n),
(k,l){\in}W(N)\right)},
\end{eqnarray}
\begin{eqnarray}
W(n)
\equiv
{\Big \{}(k,l){\Big |}
k,l=-{\lfloor}{\frac{n-1}{2}}{\rfloor},
-{\lfloor}{\frac{n-1}{2}}{\rfloor}+1,{\cdots},
+{\lfloor}{\frac{n}{2}}{\rfloor}{\Big \}},
\end{eqnarray}
\begin{eqnarray}
& &
{\langle}k,l|{\bm{U}}(n)|x,y{\rangle} \equiv 
{\frac{1}{n}}
{\exp}{\left(-i{\frac{2{\pi}kx}{n}}(kx+ly)\right)}
\nonumber\\
& &
{\hspace{1.0cm}}
{\left((x,y){\in}V(n),(k,l){\in}W(n)\right)},
\label{UnitaryMatrix}
\end{eqnarray}
and ${\bm{U}}^{\dag}(n)\equiv{\overline{{\bm{U}}(n)}}^{\rm{T}}$ 
is the adjoint matrix of ${\bm{U}}(n)$.
We assume that the prior probability density function of ${\bm{f'}}$
is given as 
\begin{eqnarray}
& &{\hspace{-1.0cm}}
P({\bm{f'}}|{\alpha},{\gamma})
{\propto}
{\exp}{\Bigg (}
-{\frac{1}{2}}
{\sum_{(k,l){\in}W(n)}}
{\bm{f'}}^{\rm{T}}{\bm{U}}^{\dag}(n)
{\left({\frac{n}{N}}\right)}^{-{\psi}}
|k,l{\rangle}
\nonumber\\
& &{\hspace{0.750cm}}
{\times}
{\left({\gamma}
+{\alpha}
{\lambda}{\left(
{\frac{2{\pi}k}{N}}
,{\frac{2{\pi}l}{N}}\right)}
\right)}
{\langle}k,l|
{\left({\frac{n}{N}}\right)}^{-{\psi}}
{\bm{U}}(n){\bm{f'}}
{\Bigg )},
\label{GaussianGraphicalModel-Prior-MSRG}
\end{eqnarray}
\begin{eqnarray}
{\lambda}{\left(
{\frac{2{\pi}k}{N}}
,{\frac{2{\pi}l}{N}}\right)}
\equiv
4-2{\cos}{\left({\frac{2{\pi}k}{N}}\right)}
 -2{\cos}{\left({\frac{2{\pi}l}{N}}\right)}.
\label{LatticeGreenFunction-GaussianGraphicalModel}
\end{eqnarray}
Moreover, we introduce the conditional probability density function
of the degraded image 
${\bm{g'}}\equiv{\left(g'_{x,y}{\big |}
(x,y){\in}V(n)\right)}$,
where the original image ${\bm{f'}}$ given 
in the reduced space $V(n)$
is assumed to satisfy
\begin{eqnarray}
& &{\hspace{-1.0cm}}
P({\bm{g'}}|{\bm{f'}},{\beta})
 = 
{\sqrt{
{\left({\frac{{\beta}}{2{\pi}}}\right)}^{|V(n)|}
}}
\nonumber\\
& &
{\times}
{\exp}{\Bigg(}
-{\frac{1}{2}}{\beta}
{\sum_{(k,l){\in}W(n)}}
{\left(
{\left({\frac{N}{n}}\right)}^{2{\phi}}
{\bm{g'}}-
{\left({\frac{N}{n}}\right)}^{2{\psi}}
{\bm{f'}}\right)}^{\rm{T}}
{\bm{U}}^{\dag}(n)|k,l{\rangle}
\nonumber\\
& &{\hspace{1.0cm}}
{\times}
{\langle}k,l|{\bm{U}}(n)
{\left(
{\left({\frac{N}{n}}\right)}^{2{\phi}}{\bm{g'}}
-
{\left({\frac{N}{n}}\right)}^{2{\psi}}{\bm{f'}}
\right)}
{\Bigg)}.
\label{GaussianGraphicalModel-DataDominant-MSRG}
\end{eqnarray}
Under these assumptions,
the marginal likelihood 
$P({\bm{g'}}|{\alpha},{\beta},{\gamma})$
in the space $V(n)$ is defined by
\begin{eqnarray}
P({\bm{g'}}|{\alpha},{\beta},{\gamma})
\equiv 
{\int}
P({\bm{g'}}|{\bm{f'}},{\beta})
P({\bm{f'}}|{\alpha},{\gamma})
d{\bm{f'}}
\label{GaussianGraphicalModel-MarginalLikelihood-MSRG}
\end{eqnarray}
Our proposed framework is designed to achieve the estimation of 
the hyperparameters ${\alpha}$, ${\beta}$ and ${\gamma}$
by maximizing 
$P{\left({\bm{g'}}
={\left({\frac{n}{N}}\right)}^{{\phi}}
{\bm{U}}^{\dag}(n){\bm{B}}(n,N){\bm{U}}(N){\bm{g}}
{\Big |}{\alpha},{\beta},{\gamma}\right)}$
when the data ${\bm{g}}$ is given.

In the present Bayesian inference method,
we assume that the data vectors ${\bm{g}}$ are generated 
from the conditional probability density function 
$P({\bm{g}}|{\bm{f}},{\beta}^{*})$ 
in Eq.(\ref{GaussianGraphicalModel-DataDominant}),
under the assumption that a parameter vector ${\bm{f}}$ is given.
By considering the average of 
the logarithm of the renormalized marginal likelihood 
${\ln}{\left(P{\left({\bm{g'}}=
{\left({\frac{n}{N}}\right)}^{{\phi}}
{\bm{U}}^{\dag}(n){\bm{B}}(n,N){\bm{U}}(N){\bm{g}}
{\Big |}{\alpha},{\beta},{\gamma}\right)}\right)}$
in the probability density function 
$P({\bm{g}}|{\bm{f}},{\beta}^{*})$, 
we can estimate the performance of our hyperparameter estimations 
using a momentum-space renormalization group approach as follows:
\begin{eqnarray}
& &{\hspace{-1.0cm}}
L_{n}{\left({\alpha},{\beta},{\gamma}{\big |}
{\beta}^{*},{\bm{f}}^{*}\right)}
\equiv
{\frac{1}{n^{2}}}
{\int}
P({\bm{g}}|{\bm{f}}={\bm{f}}^{*},{\beta}^{*})
\nonumber\\
& &
{\times}
{\ln}{\left(
P{\left({\bm{g'}}=
{\left({\frac{n}{N}}\right)}^{{\phi}}
{\bm{U}}^{\dag}(n){\bm{B}}(n,N){\bm{U}}(N){\bm{g}}
{\Big |}{\alpha},{\beta},{\gamma}\right)}
\right)}
d{\bm{g}}
\nonumber\\
& &{\hspace{-1.0cm}}
=
{\ln}{\left(
{\left({\frac{n}{N}}\right)}^{\phi}
\right)}
-{\frac{1}{2}}
{\ln}{\left({\frac{2{\pi}}{{\beta}}}\right)}
% \nonumber\\
% & &
+{\frac{1}{2n^{2}}}
{\sum_{(k,l){\in}W(n)}}
{\ln}{\left(
{\frac{
{\gamma}+{\alpha}
{\lambda}{\left({\frac{2{\pi}k}{N}},{\frac{2{\pi}l}{N}}\right)}
}
{
{\beta}+{\gamma}
+{\alpha}
{\lambda}{\left({\frac{2{\pi}k}{N}},{\frac{2{\pi}l}{N}}\right)}
}}
\right)}
\nonumber\\
& &{\hspace{-1.0cm}}
-{\frac{1}{2n^{2}}}
{\sum_{(k,l){\in}W(n)}}
{\frac{
{\beta}
{\left({\gamma}
+{\alpha}
{\lambda}{\left({\frac{2{\pi}k}{N}},{\frac{2{\pi}l}{N}}\right)}
\right)}
}
{
{\beta}^{*}
{\left(
{\beta}+{\gamma}
+{\alpha}
{\lambda}{\left({\frac{2{\pi}k}{N}},{\frac{2{\pi}l}{N}}\right)}
\right)}
}}
\nonumber\\
& &{\hspace{-1.0cm}}
-{\frac{1}{2n^{2}}}
{\sum_{(k,l){\in}W(n)}}
({\bm{f}}^{*})^{\rm{T}}{\bm{U}}^{\dag}(N){\bm{B}}^{\rm{T}}(n,N)|k,l{\rangle}
\nonumber\\
& &
{\hspace{0.50cm}}
{\times}
{\frac{
{\beta}
{\left({\gamma}
+{\alpha}
{\lambda}{\left({\frac{2{\pi}k}{N}},{\frac{2{\pi}l}{N}}\right)}
\right)}
}
{
{\beta}+{\gamma}
+{\alpha}
{\lambda}{\left({\frac{2{\pi}k}{N}},{\frac{2{\pi}l}{N}}\right)}
}}
{\langle}k,l|{\bm{B}}(n,N){\bm{U}}(N){\bm{f}}^{*}.
\label{StatisticalAverageF_LogMarginalLikelihoodB}
\end{eqnarray}
The statistical averages for the estimates of the hyperparameters
are given as
\begin{eqnarray}
& &{\hspace{-1.0cm}}
{\left({\overline{\alpha}}_{n},
{\overline{\beta}}_{n},
{\overline{\gamma}}_{n}\right)}
=
{\arg}{\max_{({\alpha},{\beta},{\gamma})}}
L_{n}{\left({\alpha},{\beta},{\gamma}{\big |}
{\beta}^{*},{\bm{f}}^{*}\right)}.
\label{StatisticalAverageF_MaxLogMarginalLikelihoodB}
\end{eqnarray}
We remark that 
the maximization of 
$L_{n}{\left({\alpha},{\beta},{\gamma}{\big |}
{\beta}^{*},{\bm{f}}^{*}\right)}$
with respect to ${\alpha}$, ${\beta}$ and ${\gamma}$
does not depend on 
the exponents ${\phi}$ and ${\psi}$ in our scale transformations
in Eqs.(\ref{ScaleTransformationF}) and (\ref{ScaleTransformationG}).
The statistical performance of our proposed scheme
is given by
\begin{eqnarray}
& &{\hspace{-1.0cm}}
D_{n}{\left({\beta}^{*},{\bm{f}}^{*}\right)}
\equiv
{\frac{1}{N^{2}}}
{\int}
||
{\bm{f}}-
{\bm{\widehat{f}}}({\overline{\alpha}}_{n},
{\overline{\beta}}_{n},
{\overline{\gamma}}_{n}|{\bm{g}})||^{2}
P({\bm{g}}|{\bm{f}}={\bm{f}}^{*},{\beta}^{*})
d{\bm{g}}
\nonumber\\
& &{\hspace{-1.0cm}}
 = 
{\frac{1}{N^{2}}}
{\sum_{(k,l){\in}W(N)}}
{\frac{1}{{\beta}^{*}}}
{\left(
{\frac{
{\overline{\beta}}_{n}
}{
{\overline{\beta}}_{n}
+{\overline{\gamma}}_{n}
+{\overline{\alpha}}_{n}
 {\lambda}{\left({\frac{2{\pi}k}{N}},{\frac{2{\pi}l}{N}}
\right)}}}
\right)}^{2}
\nonumber\\
& &{\hspace{-1.00cm}}
+
{\frac{1}{N^{2}}}
{\sum_{(k,l){\in}W(N)}}
({\bm{f}}^{*})^{\rm{T}}|k,l{\rangle}
{\left(
{\frac{{\overline{\gamma}}_{n}
       +{\overline{\alpha}}_{n}
        {\lambda}{\left({\frac{2{\pi}k}{N}},{\frac{2{\pi}l}{N}}\right)}
      }
      {{\overline{\beta}}_{n}
       +{\overline{\gamma}}_{n}
       +{\overline{\alpha}}_{n}
        {\lambda}{\left({\frac{2{\pi}k}{N}},{\frac{2{\pi}l}{N}}\right)}
      }}
\right)}^{2}
{\langle}k,l|{\bm{f}}^{*},
\nonumber\\
& &
\label{StatisicalAveragePerformanceGaussianGraphicalModel}
\label{ClosedForm-StatisicalAveragePerformanceGaussianGraphicalModel}
\end{eqnarray}
where
\begin{eqnarray}
{\widehat{f}}_{x,y}
{\left({\alpha},{\beta},{\gamma}{\big |}{\bm{g}}\right)}
=
{\int}
f_{x,y}P({\bm{f}}|{\bm{g}},
{\alpha},{\beta},{\gamma})
d{\bm{f}},
\label{MPM-GaussianGraphicalModel}
\end{eqnarray}
\begin{eqnarray}
P({\bm{f}}|{\bm{g}},{\alpha},{\beta},{\gamma})
\propto
P({\bm{g}}|{\bm{f}},{\beta})
P({\bm{f}}|{\alpha},{\gamma}).
\label{GaussianGraphicalModel-Posterior}
\end{eqnarray}

We show some numerical experiments in Fig.{\ref{Figure01}}. 
\begin{figure}[tb]
\begin{center}
{\bf{(a)}}{\hspace{4.0cm}}
{\bf{(b)}}
\\
{\hspace{0.250cm}}
\includegraphics[height=3.0cm]{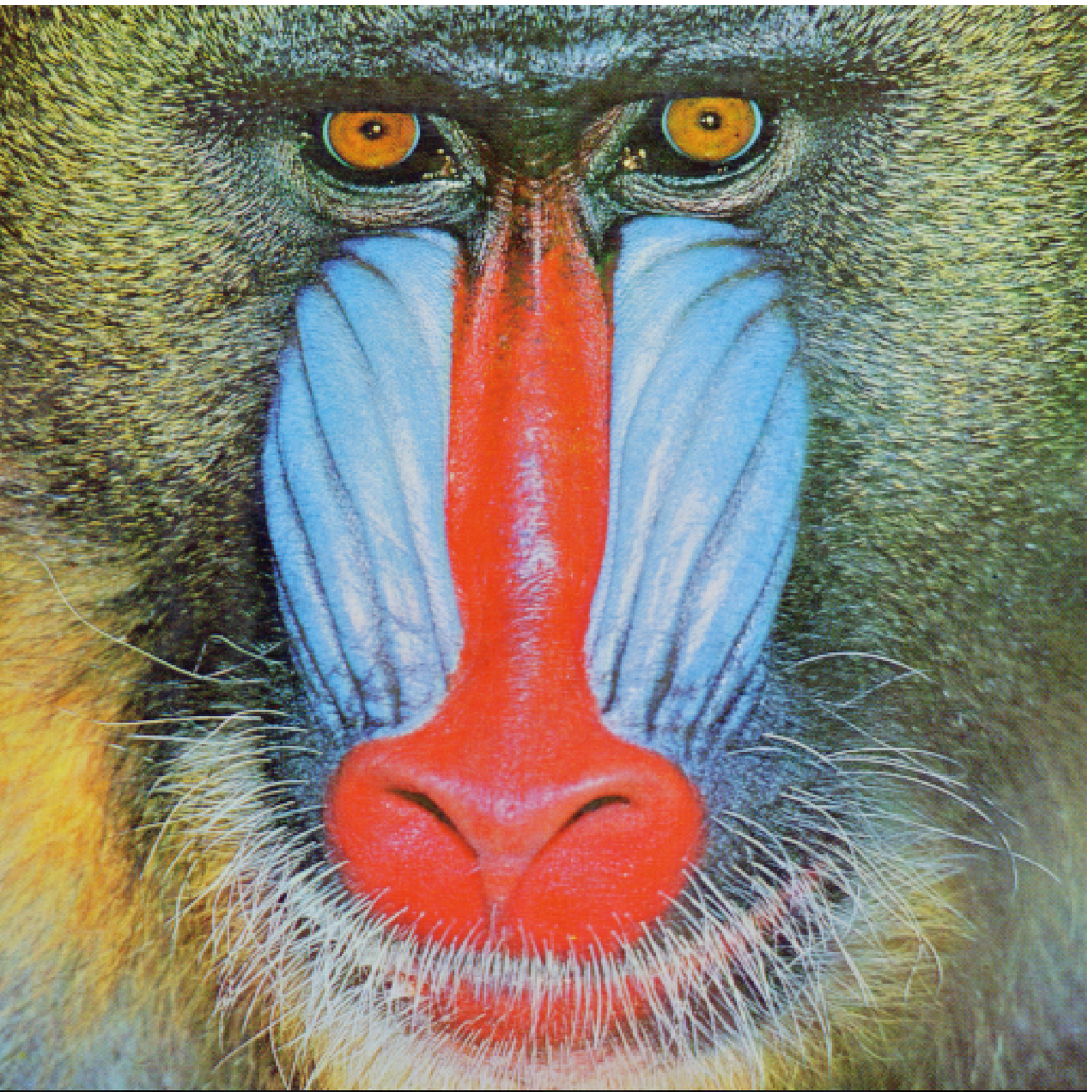}
{\hspace{0.50cm}}
\includegraphics[height=3.0cm]{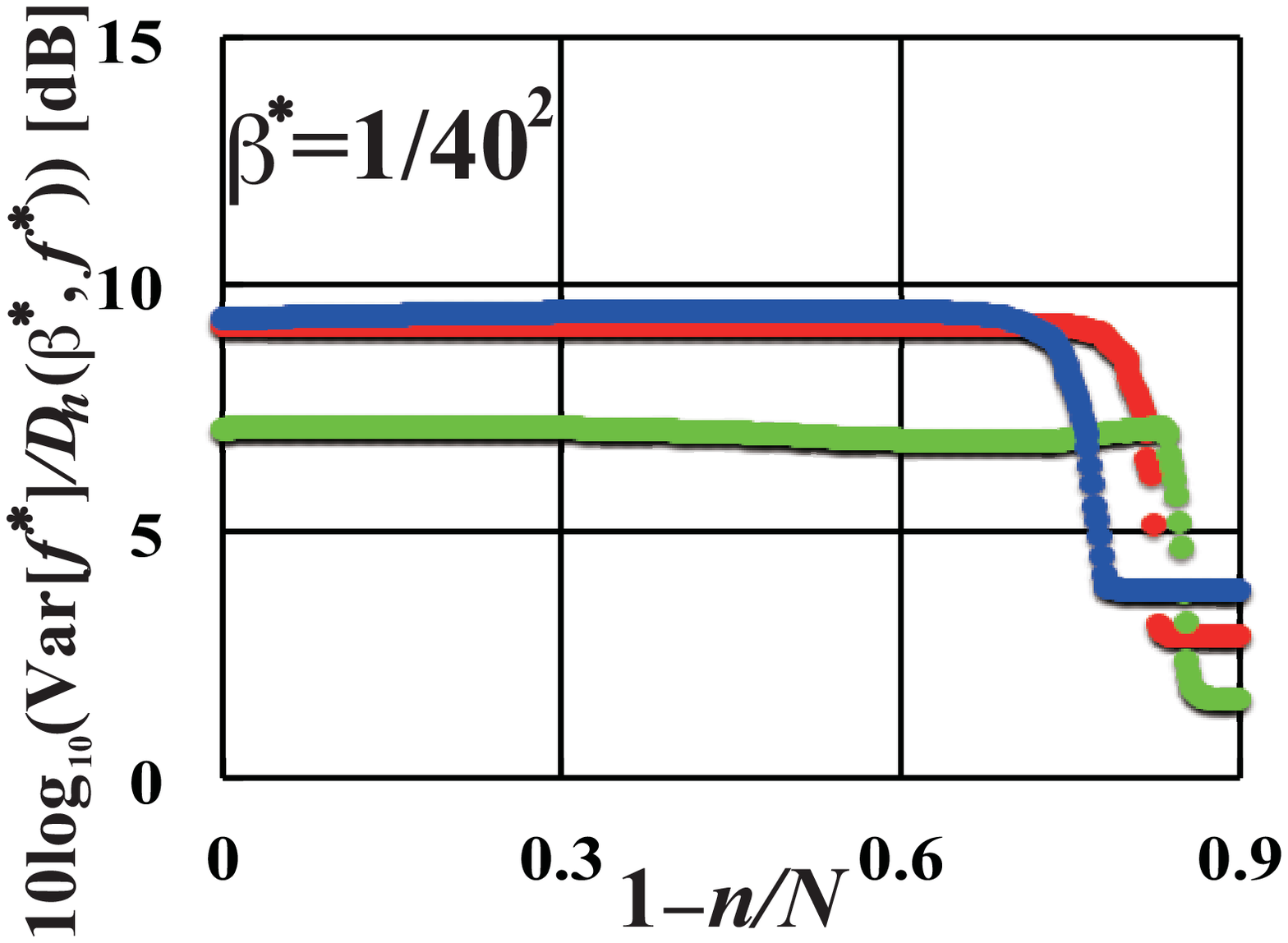}
\end{center}
\caption{
Numerical experiments for 
$D_{n}{\left({\beta}^{*},{\bm{f}}^{*}\right)}$
(a) Original image ${\bm{f}}^{*}$ ($N=512$). 
(b) Logarithm of the signal to noise ratio 
$10{\log}_{10}{\left({\rm{Var}}[{\bm{f}}^{*}]/
D_{n}{\left({\beta}^{*},{\bm{f}}^{*}\right)}\right)}$ [dB].
We set the original image ${\bm{f}}^{*}$ to 
each color intensity set of the image in Fig.{\ref{Figure01}}(a). 
${\rm{Var}}[{\bm{f}}^{*}]$ is the variance of ${\bm{f}}$.
We set the variance of additive white Gaussian noise to $40^{2}$, 
such that ${\beta}^{*}=1/40^2$.
The red, green and blue circles  
correspond to the results for each light color intensity.}
\label{Figure01}
\end{figure}
We set the original image ${\bm{f}}^{*}$ to 
each color intensity set of the image in Fig.{\ref{Figure01}}(a). 
In Fig.{\ref{Figure01}}(b), 
the solid circles in red, green and blue 
correspond to 
the logarithms of the signal to noise ratio 
$10{\log}_{10}{\left({\rm{Var}}[{\bm{f}}^{*}]/
D_{n}{\left({\beta}^{*},{\bm{f}}^{*}\right)}\right)}$
for each corresponding light color intensity, 
where ${\rm{Var}}[{\bm{f}}^{*}]$ is the variance of ${\bm{f}}^{*}$.
Here $D_{n}{\left({\beta}^{*},{\bm{f}}^{*}\right)}$
is computed by using 
Eq.(\ref{StatisicalAveragePerformanceGaussianGraphicalModel}).
It can be seen 
that the performance is almost unchanged
in the region $0<1-{\frac{n}{N}}<0.75$.

In the present paper, 
we have presented our formulation in the case of a square grid graph
in order to explain the momentum space renormalization group analysis 
in Bayesian modeling.
However, our formulation can be applied not only 
to Gaussian graphical models on a square grid graph 
but also to one on any random graph.
This remains a problem for one of our future works.

\section*{Acknowledgements}
This work was partly supported 
by the JST-CREST (No.JPMJCR1402)
for Japan Science and Technology Agency
and the JSPS KAKENHI Grant (No.25120009 and No.15K20870). 
% \section*{References}

\end{document}